# Towards Automated Cadastral Boundary Delineation from UAV data

**Sophie Crommelinck** [1,*], **Michael Ying Yang** [1], **Mila Koeva** [1], **Markus Gerke** [2], **Rohan Bennett** [3], **and George Vosselman** [1]

[1] Faculty of Geo-Information Science and Earth Observation (ITC), University of Twente, Enschede, the Netherlands; m.n.koeva@utwente.nl (M.K); michael.yang@utwente.nl (M.Y.Y.); george.vosselman@utwente.nl (G.V.)

[2] Institute of Geodesy und Photogrammetry, Technical University of Brunswick, Brunswick, Germany; m.gerke@tu-bs.de (M.G)

[3] Swinburne Business School, Swinburne University of Technology, Victoria, Australia; rohanbennett@swin.edu.au (R.B)

**\*** Correspondence: s.crommelinck@utwente.nl; Tel.: +31-53-489-5524

**Abstract:** Unmanned aerial vehicles (UAV) are evolving as an alternative tool to acquire land tenure data. UAVs can capture geospatial data at high quality and resolution in a cost-effective, transparent and flexible manner, from which visible land parcel boundaries, i.e., cadastral boundaries are delineable. This delineation is to no extent automated, even though physical objects automatically retrievable through image analysis methods mark a large portion of cadastral boundaries. This study proposes (i) a workflow that automatically extracts candidate cadastral boundaries from UAV orthoimages and (ii) a tool for their semi-automatic processing to delineate final cadastral boundaries. The workflow consists of two state-of-the-art computer vision methods, namely gPb contour detection and SLIC superpixels that are transferred to remote sensing in this study. The tool combines the two methods, allows a semi-automatic final delineation and is implemented as a publicly available QGIS plugin. The approach does not yet aim to provide a comparable alternative to manual cadastral mapping procedures. However, the methodological development of the tool towards this goal is developed in this paper. A study with 13 volunteers investigates the design and implementation of the approach and gathers initial qualitative as well as quantitate results. The study revealed points for improvement, which are prioritized based on the study results and which will be addressed in future work.

## 1. Introduction

Unmanned aerial vehicles (UAVs) are rapidly developing and increasingly applied in remote sensing, as they allow for multi-scale observations and fill the gap between ground based sampling and satellite based observations. Numerous application fields make use of the cost-effective, flexible and rapid acquisition system delivering othoimages, point clouds and digital surface models (DSMs) of high resolution [1-3]. Recently, the use of UAVs in land administration is expanding [4-9]: the high-resolution imagery is often used to visually detect and manually digitize cadastral boundaries. Such boundaries outline land parcels, for which additional information such as ownership and value are saved in a corresponding register. The resulting cadastral map is considered crucial for a continuous and sustainable recording of land rights [10]. Worldwide, the land rights of over 70% of the population are unrecognized, wherefore innovative, affordable, reliable, transparent, scalable and participatory tools for fit-for-purpose and responsible land administration are sought [11,12]. Automatically extracting visible cadastral boundaries from UAV data by providing a publicly available tool to edit and finalize those boundaries would meet this demand and improve current mapping procedures in terms cost, time and accuracy.

This study proposes a corresponding approach for UAV-based cadastral mapping. It is based on the assumption that a large portion of cadastral boundaries is physically manifested through objects such as hedges, fences, stone walls, tree lines, roads, walkways or waterways. Those visible boundaries bear the potential to be extractable with computer vision methods [13-15]. Such automatically extracted outlines require further (legal) adjudication that allows incorporating local knowledge from a human operator. In past work, a hypothetical generalized workflow for the automatic extraction of visible cadastral boundaries has been proposed [13]. It was derived from 89 studies that extract objects related to those manifesting cadastral boundaries from high-resolution optical sensor data. The synthesized workflow consists of image segmentation, line extraction and contour generation (Figure 1). For image segmentation, globalized probability of boundary (gPb) contour detection was found to be applicable for an initial detection of visible boundaries. However, the global optimization of the method does not allow to process large images. Therefore, the UAV data was reduced in resolution, which lead to a reduced localization quality [16]. To improve the localization quality at the locations of initially detected candidate boundaries is the aim of the proceeding workflow step: for line extraction, simple linear iterative clustering (SLIC) superpixels were found to coincide largely with object boundaries in terms of completeness and correctness [17]. For contour generation, gPb contour detection and SLIC superpixels are combined and processed in a semi-automatic tool that allows a subsequent final delineation of cadastral boundaries. This final step is described in this paper.

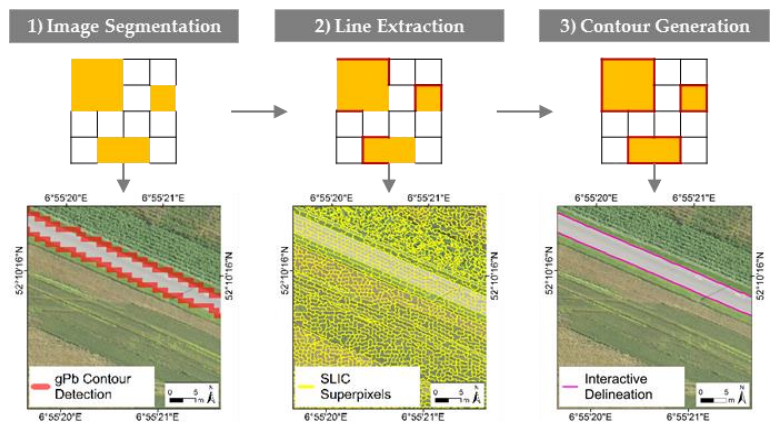

**Figure 1.** Sequence of commonly applied workflow steps proposed in [13] to extract objects related to those manifesting cadastral boundaries from high-resolution optical sensor data. For the first and second workflow step, state-of-the-art computer vision approaches have been evaluated separately and determined as efficient for UAV-based cadastral mapping [16,17]. The third workflow step as well as the overall approach is described in this paper.

*1.1. Objective and Organization of the Study*

The project that the proposed approach is developed for, seeks to offer land tenure mapping tools that align with fit-for-purpose concepts in land administration by being flexible, inclusive, participatory, affordable, reliable, attainable and upgradeable [11,18]. Within this context, this paper provides the design and implementation of a tool that addresses the lack of automation within cadastral boundary delineation. It does not aim to provide a comprehensive approach for cadastral mapping, but rather a tool for the semi-automatic extraction of visible cadastral boundaries, as well as their interactive user-guided processing to final boundaries. The study includes a subsequent delineation study that aims to evaluate the tool's design and implementation to identify and prioritize points for improvement. It further captures geometric accuracy measures compared to manually drawn reference data to quantitatively evaluate the tool's initial development phase.

The proposed approach will be used as a basis for the on-going development of a tool meeting the requirements described. Once UAV data as well as cadastral data is available through the project for places, where the concept of fit-for-purpose land administration is accepted or in place, a comprehensive comparison against current (manual) mapping procedures will be possible. To develop a tool that is comparable and in the best case superior to manual mapping procedures in terms of accuracy, processing time and usability, will be the focus of future work. Such comparisons will only make sense in areas, where a large portion of cadastral boundaries is visible, i.e., demarcated by physical objects, as the tool is designed for such cases.

The paper is structured as follows: Section 1 introduced the study in a broader context relating it to recent trends, concepts and studies, as well as to past work. In Section 2, the UAV data, each part of the methodological approach, as well as the delineation study is described. Section 3 shows results, which are discussed in Section 4. Section 5, provides concluding remarks on the entire approach and outlines unsolved issues to be addressed in future work.

## 2. Materials and Methods

### 2.1. UAV Data

For this study, an orthoimage showing a rural extent in Germany was selected. In rural areas, the portion of visible boundaries manifested through hedges, fences, stone walls, tree lines, roads, walkways or waterways is assumed to be larger compared to high-density urban areas [19]. The data was captured with indirect georeferencing, i.e., Ground Control Points (GCPs) were distributed on the ground and measured with a Global Navigation Satellite System (GNSS). The data was processed with Pix4DMapper. Table 1 shows specifications of the data capture. Figure 2 shows the orthoimage of the study area.

The reference data was manually digitized on the 0.05 m UAV orthoimage. The reference data consists of outlines of objects demarcating visible cadastral boundaries, such as roads, as well as outlines of agricultural fields, waterbodies and forest. As the tool is designed for areas, in which a large portion of cadastral boundaries is visible, i.e., demarcated by a physical object, the UAV image was chosen accordingly: most lines in the reference data demarcate visible objects.

**Table 1.** Specifications of the UAV dataset used for the delineation study (Section 2.3).

| Location | Latitude/ Longitude | UAV Model | Camera/Focal Length [mm] | Forward/Sideward Overlap [%] | GSD [m] | Extent [m] |
|---|---|---|---|---|---|---|
| Amtsvenn, Germany | 52.17335/ 6.92865 | GerMAP G180 | Ricoh GR/13.3 | 80/65 | 0.05 | 1000x1000 |

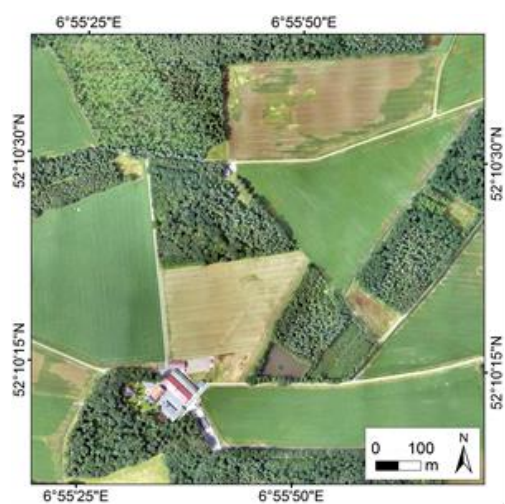

**Figure 2.** UAV orthoimages of Amtsvenn, Germany used for the delineation study (Section 2.3). All methods included in the image processing workflow (Section 2.2) have been tested and validated on further UAV orthoimages [16,17].

*2.2. Image Processing Workflow*

The image processing workflow is based on the workflow proposed in [13] and shown in Figure 1. Its three components are described separately in Sections 2.2.1-2.2.3:

2.2.1. Image Segmentation – gPb Contour Detection

Contour detection refers to finding closed boundaries between objects or segments. Globalized probability of boundary (gPb) contour detection refers to the processing pipeline of gPb-owt-ucm, which is visualized in Figure 3, explained in the following and based on [20]. The pipeline originates from computer vision and aims to find closed boundaries between objects or segments within an image. This is achieved through combining edge detection and hierarchical image segmentation, while integrating image information on texture, color and brightness on both a local and a global scale.

In a first step, oriented gradient operators for brightness, color and texture are calculated on two halves of differently scaled discs. The cues are merged based on a logistic regression classifier resulting in a posterior probability of a boundary, i.e., an edge strength per pixel. This local image information is combined trough learning techniques with global image information that is obtained through spectral clustering. The learning steps are trained on natural images from the 'Berkeley Segmentation Dataset and Benchmark' [21]. By considering image information on different scales, relevant boundaries are verified, while irrelevant ones, e.g., in textured regions, are eliminated.

In the second step, initial regions are formed from the oriented contour signal provided by a contour detector through oriented watershed transformation (owt). Subsequently, a hierarchical segmentation is performed through weighting each boundary and their agglomerative clustering to create an ultrametric contour map (ucm) that defines the hierarchical segmentation.

The overall results consists of (i) a contour map, in which each pixel is assigned a probability of being a boundary pixel, and (ii) a binary boundary map containing closed contours, in which each pixel is labeled as 'boundary' or 'no boundary'. The approach has shown to be applicable to UAV orthoimages for an initial localization of candidate object boundaries. Due to the global optimization, UAV orthoimages of extents larger than 1000 x 1000 pixels need to be reduced in resolution, in order to be processed with the original implementation. This leads to a reduced localization quality [16]. To improve the localization quality of initially detected candidate boundaries is the aim of the proceeding workflow steps.

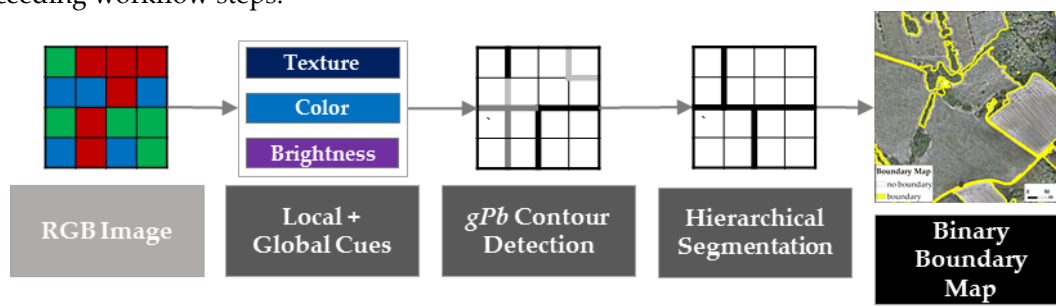

**Figure 3.** Processing pipeline of globalized probability of boundary (gPb) contour detection and hierarchical image segmentation resulting in a binary boundary map containing closed segment boundaries. The approach is described in [20]. The source code is publicly available in [22].

2.2.2. Line Extraction – SLIC Superpixels

Simple linear iterative clustering (SLIC) superpixels originate from computer vision and were introduced in [23]. Superpixels aim to group pixels into perceptually meaningful atomic regions and can therefore be located between pixel- and object-based approaches. The approach allows to compute image features for each superpixel rather than each pixel, which reduces subsequent processing tasks in complexity and computing time. Further, the boundaries of superpixels adhere well to objects outlines within the image and can therefore be used to delineate objects.

When comparing state-of-the-art superpixel approaches, SLIC superpixels have outperformed comparable approaches in terms of speed, memory efficiency, compactness and correctness of outlines [24-29]. The approach visualized in Figure 4 was introduced in [30] and extended in [24]. SLIC considers image pixels in a 5D space, in terms of their L*a*b values of the CIELAB color space and their x and y coordinates. Subsequently, the pixels are clustered based on an adapted k-means clustering. The clustering considers color similarity and spatial proximity. SLIC implementations are available in OpenCV [31], VL Feat [32], GDAL [33], Scikit [34], Matlab [35] and GRASS [36].

The approach has shown to be applicable to UAV orthoimages of 0.05 m ground sample distance (GSD). Further, cadastral boundaries demarcated through physical objects often coincide with the outlines of SLIC superpixels [17]. While [17] is based on a Matlab implementation [35], this study applies a GRASS implementation [36]. The latter appears to have a better boundary adherence for smaller superpixels. Further, it will be easier to integrate the method in the QGIS plugin of the subsequent workflow step (Section 2.2.3), once the module is transferred from being a GRASS Add-on to a regular GRASS module [37], which ensures its availability in QGIS processing [38].

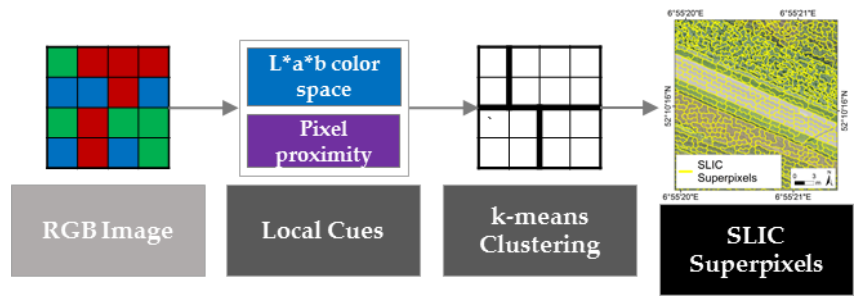

**Figure 4.** Processing pipeline of simple linear iterative clustering (SLIC) resulting in agglomerated groups of pixels, i.e., superpixels, whose boundaries outline objects within the image. The approach is described in [30] and extended in [24].

2.2.3. Contour Generation – Interactive Delineation

Contour generation refers to generating a vectorized and topologically connected network of SLIC outlines from Section 2.2.2 that surround candidate regions from Section 2.2.1. This step combines the detection quality of gPb contour detection with the localization quality of SLIC superpixels. This workflow step consists of two parts shown in Figure 5: (i) the automatic combination of gPb and SLIC and (ii) the semi-automatic delineation of final boundaries through a human operator. Both parts are implemented in a publicly available QGIS plugin [39] with a Graphical User Interface (GUI) shown in Figure 6. The source code with test data, as well as a manual including screenshots and a link to a YouTube video explaining its use are provided in [40].

In (i), only those SLIC outlines are kept, that are in proximity of gPb contour detection outlines. This is done by applying a buffer of 5 m radius around the gPb outlines. The topology of the remaining SLIC lines is cleaned before they are transferred to a network with nodes on each intersection of two or more SLIC lines. The network and nodes are displayed to the user (Figure 7).

In (ii), the user is asked to select two or more nodes along a land parcel boundary. These are then automatically connected based on a Steiner network method [41]. This method searches the shortest path along the remaining SLIC outlines between the nodes that the user selects. A sinuosity measure is calculated for the created line, in order to provide the user with an indication on the line's usability. Sinuosity measures to which extent a line between two points varies from their direct connection, i.e., the ratio between the Euclidean distance between two points and the length of the line connecting the two points. The range of the sinuosity measure is [0; 1]. It is equally divided into three parts to color the line according to a traffic light evaluation system in red, yellow and green. The line is displayed accordingly to indicate the line's usability to the user (Figure 7). Thereafter, the user has the option to simplify the created line, which is done based on the Douglas-Peucker approach [42]. This algorithm simplifies the line by creating a curve along a series of points and gradually reducing the number of points. The user further has the option to manually edit the line or specific nodes of the line by making use of the extensive QGIS editing functionalities [43]. Further options consist of deleting or accepting the line. Choosing the latter, leads to a display of the initial network and nodes and the request to select a new set of nodes to be connected.

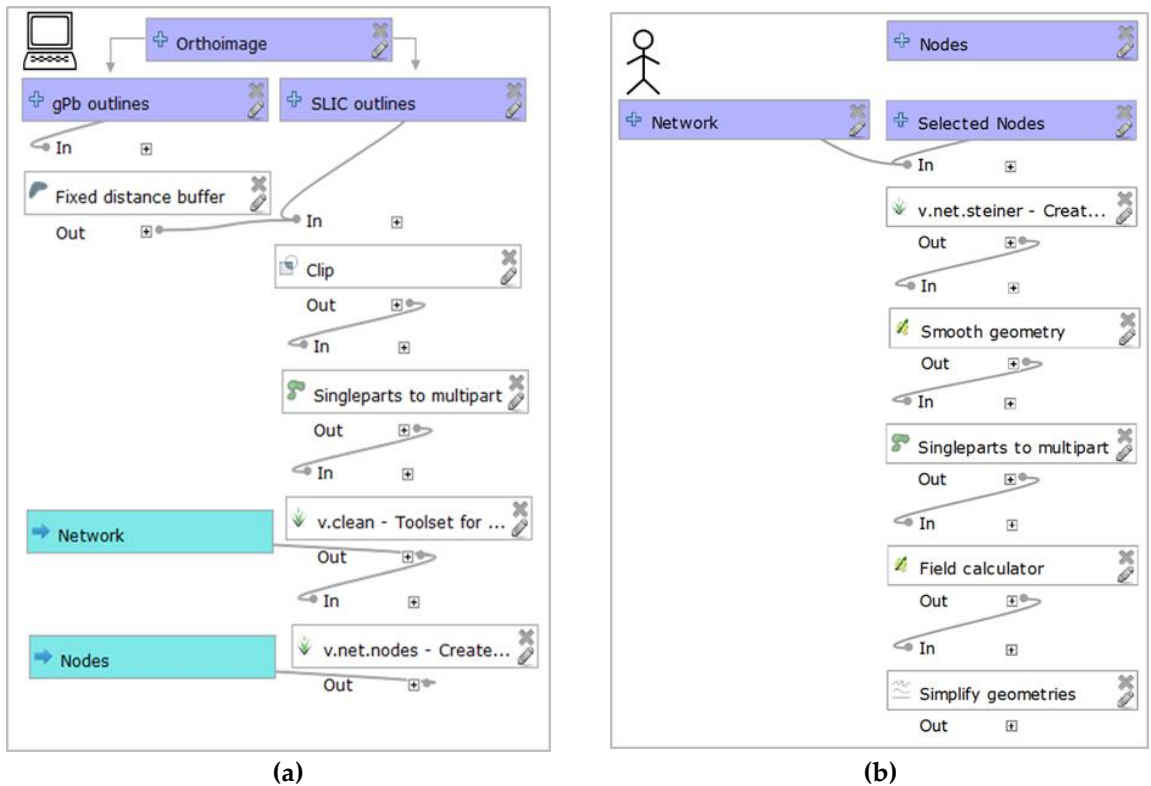

**Figure 5.** QGIS processing model showing the processing pipeline of the BoundaryDelineation QGIS plugin [39]. (**a**) shows the automatic processing of Step I in the plugin, i.e., combining gPb contour detection and SLIC superpixels derived from the UAV orthoimage. (**b**) shows the user-guided processing of step II in the plugin, i.e., connecting the created network via nodes selected by a user and optional further processing steps that the user can select from (accept, edit, simplify or delete the line). The input files per step are shown in dark blue, the output files in light blue and the processing steps with the corresponding processing modules from GRASS and QGIS are shown in white.

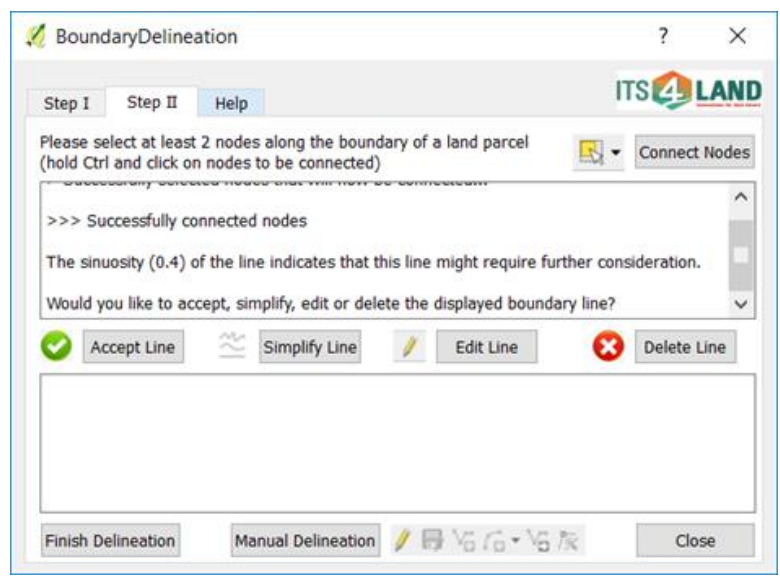

**Figure 6.** Interface of step II of the BoundaryDelineation QGIS plugin that automatically combines the results of gPb contour detection and SLIC superpixels and further supports the subsequent semi-automatic delineation of final boundaries through a human operator.

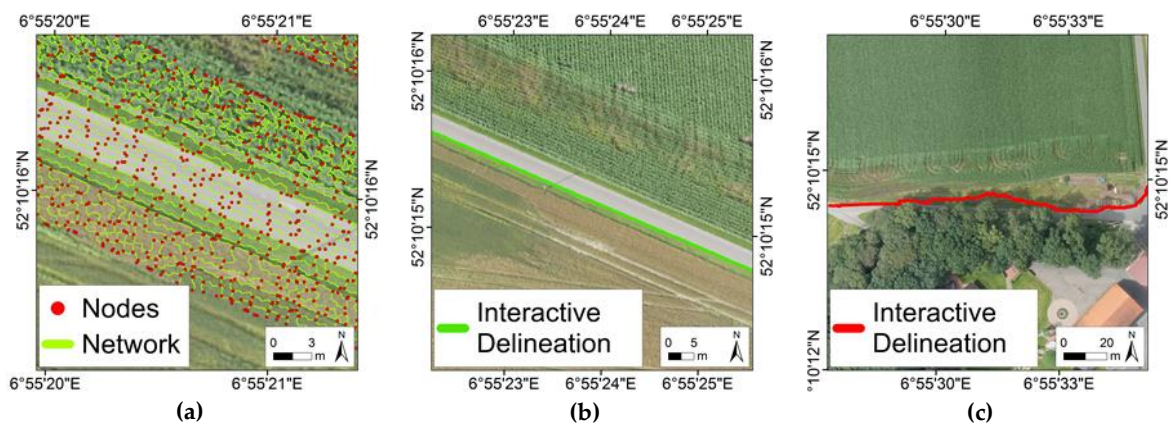

**Figure 7.** (**a**) Nodes and network generated automatically through combining the outlines of gPb contour detection and SLIC superpixels in step I of the QGIS plugin. From these data, the user is asked to select at least two nodes along a boundary that are then automatically connected along the shortest path of the network lines. The line's usability is indicated to the user based on a sinuosity measure displayed in the GUI (**Figure 6**) and by displaying the line according to a traffic light evaluation system in (**b**) green, yellow or (**c**) red.

### 2.3. Delineation Study

In order to test the design and implementation of the image processing workflow (Section 2.2), a study was conducted. This study does not consider the interactive delineation as comparable with manual delineation, but aims to create an impression of the tool's current state and to identify and prioritize points for improvement. Participants possessing at least undergraduate-level geospatial science knowledge were asked to delineate visible boundaries within the described UAV orthoimage (Section 2.1) with the developed QGIS plugin (Section 2.2.3). Before and following the task, the participants were asked to fill out a survey: the first survey [44] aimed to capture the educational background, the experience with (Q)GIS software and on-screen digitization, as well as potential working experience within these fields. The second survey [45] aimed to capture a qualitative impression on the plugin's usability, its limitations and recommendation for its improvement.

The delineation study was designed for three hours. At first, a short overview about the research was provided. The participants were asked to install the required software and download all data to their computers. As not all boundaries were visible and thereby not delineable with the plugin, manual delineation in QGIS was explained and demonstrated. After the participants had tested the manual delineation, they were asked to manually digitize visible land parcel boundaries in QGIS. Since they were lacking the local knowledge about the boundaries' location that a surveyor would usually have, the desired visible land parcel boundaries, i.e., the reference data, were projected to the wall and could be checked on one computer.

The subsequent interactive delineation followed the same procedure: it was explained and demonstrated before the participants could test it. Then, the participants were asked to delineate visible land parcel boundaries as fast and accurate as possible. They were advised to delineate visible boundaries demarcated by objects with distinct outlines with the plugin, and to delineate the remaining boundaries manually. The gPb contour detection and SLIC superpixels outlines required by the plugin were provided to the participants. After having submitted the delineated boundary files, they were asked to fill out the second questionnaire on their qualitative impression of the interactive delineation.

### 2.4. Accuracy Assessment

In order to assess the results quantitatively, the boundaries delineated by the participants were compared to the reference data in terms of localization quality. The localization quality measures to what extent the boundaries delineated by the participants deviated from the exactly localizable parts of the reference data in a geometric sense. This was done by considering exactly localizable objects only, i.e., outlines of trees or bushes were excluded from the reference data.

The interactively delineated data were buffered with radius buffer distances ranging from 0 to 1 m at increments of 0.2 m. The assessment was implemented pixel-based. The subset of the reference data was overlaid with the interactive delineation data to calculate the confusion matrix per buffer distance. Each pixel was labelled as true positive (TP), true negative (TN), false positive (FP) or false negative (FN) [46,47]. Pixels with a TP label were summated per distance to the reference data and written to the confusion matrix. This allowed to quantify the distances to which the delineated data deviated from the reference data, i.e., the accuracy of boundaries in a geometric sense.

## 3. Results

In Section 3.1, the study results are analyzed in terms of the two surveys that the participants were asked to fill out before and after conducting the delineation study. This aims to evaluate the plugin qualitatively. In Section 3.2, the study results are analyzed in terms of the delineations performed by the participants according to the described accuracy assessment (Section 2.4). This aims to evaluate the plugin quantitatively.

### 3.1. Participants Survey

The delineation study took place on 23 May 2017 with 13 participants. The participants were M.Sc. or Ph.D. students at the Faculty of Geo-Information Science and Earth Observation (ITC) of the University of Twente. The M.Sc. students were enrolled in Geo-informatics (23%), Land Administration (15%) and Applied Earth Science (15%). The Ph.D. students were doing their research in earth system analysis (15%), urban and regional planning as geo-information management (15%), earth observation science (8%) and spatial ecology (8%). The majority of participants had working experience as GIS specialist/technician/analyst, surveyor, urban planner, or in academia. Their background knowledge on using QGIS software, performing on-screen digitization, and working with GIS software is visualized in Figure 7.

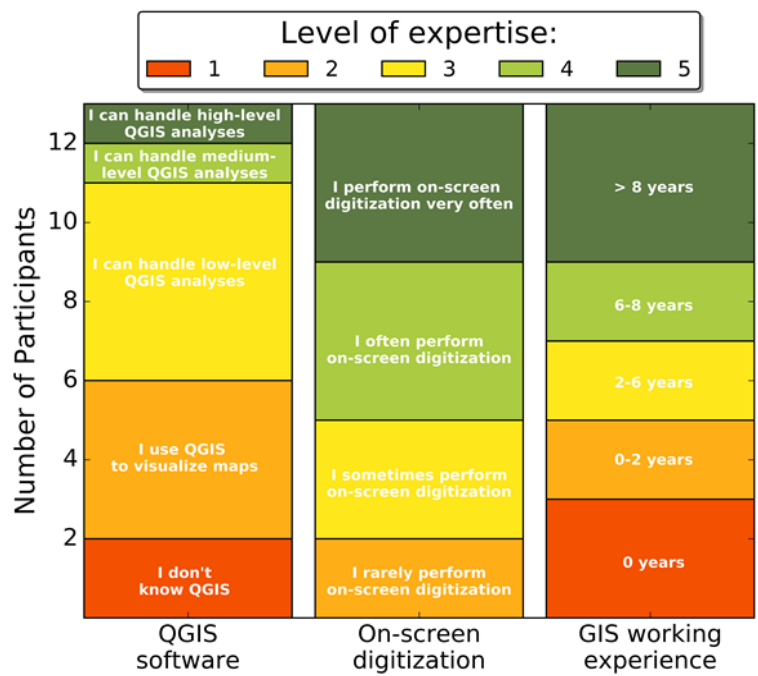

**Figure 7.** The participants' background knowledge on using QGIS software, performing on-screen digitization, and their working experience with GIS software. The participants were asked to assign themselves a level of expertise ranking from one (low) to five (high).

From the survey that the participants were asked to fill out after conducting the delineation study, the following results were obtained: the participants stated that 60% of interactively delineated lines, were directly acceptable as boundaries. For the remaining 40%, the participants applied further line simplification or editing: even though the displayed line already included a small degree of simplification, the line delineated objects with a 0.05 m accuracy. This represented the objects precisely, but not the desired boundary. The level of detail was too high. In addition, there was a large portion of imprecisely delineable boundaries, e.g., along forest or agricultural fields. Those lines were manually delineated with the QGIS editing functionalities, to which the participants were directed via the GUI of the plugin.

For the following results, the frequency with which each issue was raised decreases from the beginning to the end of the list. As main reasons for the plugin's inability to delineate land parcel boundaries, the following reasons were identified: (i) the land parcel boundaries were not visible, e.g., covered by trees, or land parcel boundaries were fuzzy/irregular, e.g. at the outlines of sandy roads, (ii) plugin implementation errors, e.g. not-working of automatic renaming of layers or inability to connect large number of nodes, and (iii) QGIS software instability. In cases of (i), the participants were instructed to delineate invisible boundaries according to the reference data.

As main recommendations to facilitate the use of the plugin the following points were listed: (i) increased percentage of visible boundaries in the UAV orthoimage, (ii) longer experience in working with the plugin, (iii) improved usability of the plugin, and (iv) more knowledge about the boundaries' exact location.

According to the participants, the plugin bears most potential when applied for the delineation of cadastral boundaries in areas, where cadastral boundaries are clearly visible, with no vegetation covering the boundaries. The listed potential areas of successful application included rural areas, e.g., for agricultural field delineation, as well as urban areas, e.g. for, building delineation. Some participants having a local African surveyor background and being aware of prevailing limitations in terms of resources and capacities saw potential in using tools of the proposed kind.

The following general recommendations to improve the plugin were listed by the participants: (i) separating the tasks of interactive delineation and manual editing more comprehensively in the GUI, (ii) docking the plugin to the QGIS toolbar to improve usability, (iii) decreasing the computing time for the creation of the node and network files (step I of the plugin), which took up to 10 minutes, (iv) decreasing the number of proposed nodes to select from, (v) allowing to simplify created lines more comprehensively, (vi) allowing to connect larger number of nodes, (vii) automatically enabling snapping in QGIS when plugin is opened, (viii) automatically keeping the last node of the previously created line selected, and (ix) including further topological checks of a created line besides its sinuosity.

In comparison to the manual delineation, the participants stated that they experienced the manual delineation as being easier and faster. Some participants favored not being restricted to select from a set of predefined nodes to delineate boundaries. The confidence in the localization quality of lines that were clearly visible, e.g., the outlines of roads, was high for the interactive lines, as those were delineated precisely by the plugin. Further, it was stated as an advantage that the plugin suggested, where to start delineating, which saved time in comparison to an entirely manual delineation.

*3.2. Accuracy Assesment*

For the interactive delineation, 71% of TP pixels are located within 0 - 0.2 m distance, 16% within 0.21 - 0.4 m distance, 7% within 0.41 - 0.6 m distance and 7% within 0.61- 1 m distance of the reference data (Figure 8).

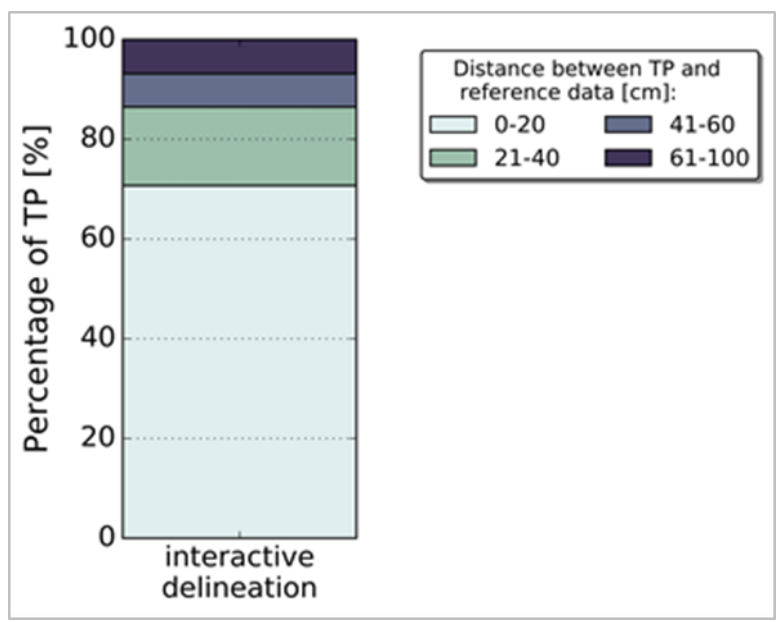

**Figure 8.** Accuracy assessment for the interactive delineation in terms of the localization quality, which shows the percentages of True Positive (TP) pixels falling within a certain distance to the reference data.

## 4. Discussion

Overall, the participants, of which a majority had experience in surveying or geospatial sciences, valued the interactive delineation for its suggestion of areas to search for candidate boundaries. A complex UAV orthoimage with visible and invisible boundaries was chosen to evaluate the tool's applicability to different scenarios: the tool performed best in cases, where a large number of cadastral boundaries was visible, as the workflow results in outlines of physical objects localizable in the image. Especially the outlines of objects with an exactly localizable boundary, such as building and road outlines, were precisely delineable with the plugin.

The interactive delineation showed a high localization quality, since it is based on SLIC outlines that take into account the 0.05 m GSD of the input UAV orthoimage. Subsequently, the interactive lines accurately run along the outlines of the objects within the image that delineate cadastral boundaries. However, there were cases, where such objects were not available or the boundaries were not visible. In such cases, the participants manually delineated the boundaries. Even though they were asked to delineate according to the reference data, they often delineated according to own boundary conceptions. Further, the balance between time and accuracy was conceived differently among the participants. The study therefore does not allow to make representative statements on the required time or the completeness and correctness of delineated boundaries. However, the study helped to collect and prioritize points on facilitating the required user interaction.

In future work, the approach will be improved in terms of the points raised by the participants. Those points will be discussed with QGIS developers and improved accordingly. This includes the tools' usability and functionalities, i.e., a reduction of proposed nodes and a more comprehensive line simplification. Furthermore, the workflow could be applied to DSMs, instead of orthoimages only, as proposed for SLIC superpixels in [48]. In order to reduce the amount of network lines and nodes to the user, both could be attributed with weights representing information on local color gradients and centrality measures. This is done similarly in [49] to extract superpixels belonging to roads. Network lines and nodes with low values, could be considered as less important and not displayed to the user.

To thoroughly evaluate the tools applicability for fit-for-purpose cadastral mapping, the tool should be evaluated in terms of the concept's criteria being innovative, affordable, reliable, transparent, scalable and participatory [11]. This could be done, once the tool design and implementation is fully developed. Such an evaluation could compare current mapping procedures (e.g., manual digitization from optical sensor data) versus the developed interactive delineation. For such an evaluation, the following aspects will be considered: (i) use of multiple UAV-images of different areas, (ii) capture of portion of cadastral boundaries being visible per area, (iii) capture of precise processing time per workflow step and (iv) definition of locally required outlining accuracy to define the buffer size for accuracy assessment measures. The listed points appear to be crucial to retrieve representative quantitative results considering time and accuracy. Instead of asking volunteer students to conduct the study, it might be advisable to ask local surveyors with working experience in cadastral mapping. Reducing the number of participants would allow to guide and supervise the adherence to study instructions more precisely. The accuracy measures could then be extended to capture the detection quality, i.e., the errors of commission and omission. Such a study could be done on its4land use case locations in Rwanda and Kenya.

## 5. Conclusion

This study contributes to the recent trend of improving fit-for-purpose cadastral mapping. This is done by proposing a tool that automates and facilitates the delineation of visible cadastral boundaries: the tool combines image analysis methods, namely globalized probability of boundaries (gPb) contour detection and simple linear iterative clustering (SLIC) superpixels. Their combination results in an initial set of network lines and nodes from which a user generates final cadastral boundary lines with the help of a publicly available QGIS plugin [39]. The approach is designed for areas, in which object contours are clearly visible and coincide with cadastral boundaries.

Volunteers were asked to test the entire workflow: the study revealed points for improvement on the design and implementation of the tool dealing mostly with the users' interaction. These points will be addressed in future work before testing the tool in real-world scenarios against manual cadastral mapping. The final tool is intended to support cadastral mapping in areas, where a large portion of cadastral boundaries is visible, and where fit-for-purpose land administration is accepted or in place.

**Acknowledgments:** This work was supported by its4land, which is part of the Horizon 2020 program of the European Union [project number 687828]. We are grateful to Claudia Stöcker for capturing, processing and providing the UAV data.

**Author Contributions:** Sophie Crommelinck has mainly performed this research, as she conducted the delineation study, including the preparation and processing of the data. Michael Ying Yang, Mila Koeva, Markus Gerke, Rohan Bennett and George Vosselman contributed to the design, analysis and interpretation of the study. Sophie Crommelinck wrote the manuscript with contributions from Michael Ying Yang, Mila Koeva, Markus Gerke, Rohan Bennett and George Vosselman.

**Conflicts of Interest:** The authors declare no conflict of interest. The founding sponsors had no role in the design of the study; in the collection, analyses, or interpretation of data; in the writing of the manuscript, and in the decision to publish the results.

12 of 146. Mumbone, M.; Bennett, R.; Gerke, M.; Volkmann, W. In *Innovations in boundary mapping: Namibia, customary lands and UAVs,* World Bank Conference on Land and Poverty, Washington DC, USA, 23-27 March, 2015; pp 1-22.
7. Volkmann, W.; Barnes, G. In *Virtual surveying: Mapping and modeling cadastral boundaries using Unmanned Aerial Systems (UAS),* FIG Congress: Engaging the Challenges - Enhancing the Relevance, Kuala Lumpur, Malaysia, 16-21 June, 2014; pp 1-13.
8. Maurice, M.J.; Koeva, M.N.; Gerke, M.; Nex, F.; Gevaert, C. In *A photogrammetric approach for map updating using UAV in Rwanda,* GeoTechRwanda, Kigali, Rwanda, 18-20 November, 2015; pp 1-8.
9. Jazayeri, I.; Rajabifard, A.; Kalantari, M. A geometric and semantic evaluation of 3D data sourcing methods for land and property information. *Land Use Policy* **2014**, *36*, 219-230.
10. Williamson, I.; Enemark, S.; Wallace, J.; Rajabifard, A. *Land administration for sustainable development*. ESRI Press Academic: Redlands, CA, USA, 2010; p 472.
11. Enemark, S.; Bell, K.C.; Lemmen, C.; McLaren, R. *Fit-For-Purpose land administration*. International Federation of Surveyors: Frederiksberg, Denmark, 2014; p 42.
12. Zevenbergen, J.; De Vries, W.; Bennett, R.M. *Advances in responsible land administration*. CRC Press: Padstow, UK, 2015; p 279.
13. Crommelinck, S.; Bennett, R.; Gerke, M.; Nex, F.; Yang, M.; Vosselman, G. Review of automatic feature extraction from high-resolution optical sensor data for UAV-based cadastral mapping. *Remote Sensing* **2016**, *8*, 1-28.
14. Zevenbergen, J.; Bennett, R. In *The visible boundary: More than just a line between coordinates,* GeoTechRwanda, Kigali, Rwanda, 18-20 November, 2015; pp 1-4.
15. Bennett, R.; Kitchingman, A.; Leach, J. On the nature and utility of natural boundaries for land and marine administration. *Land Use policy* **2010**, *27*, 772-779.
16. Crommelinck, S.; Bennett, R.; Gerke, M.; Yang, M.; Vosselman, G. Contour detection for UAV-based cadastral mapping. *Remote Sensing* **2017**, *9*, 1-13.
17. Crommelinck, S.; Bennett, R.; Gerke, M.; Koeva, M.; Yang, M.Y.; Vosselman, G. In *SLIC Superpixels for Object Delineation from UAV Data,* UAV-g, Bonn, Germany, 04-07 September, 2017; pp 1-8 (accepted).
18. Rohan Bennett, M.G., Joep Crompvoets, Serene Ho, Angela Schwering, Malumbo Chipofya, Carl Schultz, Tarek Zein, Mireille Biraro, Berhanu Alemie, Robert Wayumba, Kaspar Kundert, Sophie Crommelinck, Claudia Stöcker In *Building Third Generation Land Tools: Its4land, Smart Sketchmaps, UAVs, Automatic Feature Extraction, and the GeoCloud,* Annual World Bank Conference on Land and Poverty (Responsible Land Governance: Towards and Evidence Based Approach), Washington D.C. (US), 20-24 March 2017, 2017; pp 1-23.
19. Kohli, D.; Bennett, R.; Lemmen, C.; Asiama, K.; Zevenbergen, J. In *A Quantitative Comparison of Completely Visible Cadastral Parcels Using Satellite Images: A Step towards Automation,* FIG Working Week 2017 (Surveying the world of tomorrow - From digitalisation to augmented reality), Helsinki, Finland, May 29-June 2, 2017.
20. Arbelaez, P.; Maire, M.; Fowlkes, C.; Malik, J. Contour detection and hierarchical image segmentation. *Pattern Analysis and Machine Intelligence* **2011**, *33*, 898-916.
21. Arbeláez, P.; Fowlkes, C.; Martin, D. Berkeley segmentation dataset and benchmark. Available online: https://www2.eecs.berkeley.edu/Research/Projects/CS/vision/bsds/ (accessed on 10 November 2016).